\documentclass[letterpaper, 10 pt, conference]{ieeeconf}
\IEEEoverridecommandlockouts
\usepackage{graphicx, cite, amsfonts, amsmath, amssymb, tabularx, booktabs, mathtools, float, empheq, breqn, mdframed, subcaption, siunitx}
\usepackage{pifont}
\usepackage[ruled,vlined,linesnumbered]{algorithm2e}
\title{\LARGE \bf VSL-Skin: Individually Addressable Phase-Change Voxel Skin for Variable-Stiffness and Virtual Joints Bridging Soft and Rigid Robots}
\author{
Zihan Zeng$^{1}$,
Jiajun An$^{2}$,
Preston Luk$^{3}$,
and Upinder Kaur$^{2,*}$%
\thanks{$^{1}$School of Mechanical Engineering, Purdue University, West Lafayette, IN, USA.
$^{2}$Department of Agricultural and Biological Engineering, Purdue University, West Lafayette, IN, USA.
$^{3}$School of Aeronautics and Astronautics, Purdue University, West Lafayette, IN, USA.
$^{*}$Corresponding author: kauru@purdue.edu.}
}
\begin{document}
\maketitle
\thispagestyle{plain}
\pagestyle{plain}
\begin{abstract}
Soft robots exhibit compliance but lack load support and pose retention, while rigid robots provide structural capacity but sacrifice adaptability. Existing variable-stiffness approaches operate at segment or patch scales, preventing precise spatial control over stiffness distribution and virtual joint placement. This paper presents the Variable Stiffness Lattice Skin (VSL-Skin), the first system enabling individually addressable voxel-level morphological control with centimeter-scale precision. The system achieves three capabilities: nearly two orders of magnitude stiffness modulation across axial (\(\mathbf{15-1200}\) N/mm), shear (\(\mathbf{45-850}\) N/mm), bending (\(\mathbf{8\times10^2-3\times10^4}\) N/deg), and torsional modes with centimeter-scale spatial control; the first demonstrated 30\% axial compression in phase-change systems while maintaining structural integrity; and autonomous component-level self-repair through thermal cycling that eliminates fatigue accumulation and enables programmable sacrificial joints for predictable failure management. Selective voxel activation creates six canonical virtual joint types with programmable compliance while preserving structural integrity in non-activated regions. The platform incorporates closed-form design models and finite element analysis for predictive synthesis of stiffness patterns and joint placement. Experimental validation demonstrates 30\% axial contraction, thermal switching in 75 second cycles, and cut-to-fit integration that preserves addressability after trimming. The row-column architecture enables platform-agnostic deployment across diverse robotic systems without specialized infrastructure. This framework establishes morphological intelligence as an engineerable system property, fundamentally advancing autonomous reconfigurable robotics.
\end{abstract}

\section{INTRODUCTION}

Rigid links and discrete joint morphologies in conventional robots enable precision and high payload capacity but impair adaptability in unstructured environments~\cite{cianchetti_2014_soft, manti_2016_stiffening}. Soft robots offer compliance for safe interaction and environmental adaptation, yet lack the structural rigidity required for manipulation tasks and pose retention~\cite{rus_2015_design,bruder_2023_increasing,oncayyasa_2023_an}. This fundamental compliance–rigidity trade-off constrains performance across domains and motivates autonomous morphological frameworks that dynamically tune stiffness to match task demands.

Variable stiffness addresses this dichotomy by enabling robots to adjust mechanical properties \emph{in situ} autonomously. Biological examples demonstrate the value of reversible, spatially resolved stiffness modulation: elephant trunks selectively stiffen to grasp heavy objects while maintaining flexibility; octopus arms generate transient joints to improve reach and manipulation~\cite{kier_1985_tongues}; echinoderms exploit mutable collagenous tissues to control body stiffness for locomotion and defense~\cite{candia_2024_mutable}. These systems achieve superior performance through localized and reversible stiffness control, enabling both compliance and load support.

Existing variable-stiffness approaches have demonstrated significant stiffness modulation through diverse mechanisms. Granular jamming systems confine loose particles to achieve reversible stiffening in grippers, continuum arms, and wearable devices~\cite{hauser_2017_jammjoint, brown_2010_universal, liu_2021_a, narang_2018_mechanically}. Layer jamming stacks flexible sheets with tunable interlayer friction, enabling stiffness control in manipulators and surgical tools~\cite{wang_2019_electrostatic, kim_2013_a, choi_2019_soft, kim_2012_design, song_2025_a, an_2025_bioinspired, mitsuda_2017_variablestiffness, chenal_2014_variable}. Antagonistic actuation co-contracts opposing actuators to increase structural rigidity~\cite{shiva_2016_tendonbased, gao_2023_programmable, pardomuan_2024_vabricbeads}. Material-based solutions employ phase-change alloys and stimuli-responsive polymers for intrinsic modulus control~\cite{song_2025_a, zhang_2024_design, shamsaalharthy_2024_variable}. Robotic skin approaches externalize these mechanisms into conformable sheets for platform-independent deployment~\cite{booth_2018_omniskins, shah_2023_robotic, gao_2023_programmable, shah_2020_jamming}. However, despite these advances, most systems implement stiffness changes at the level of whole segments or patches and are embedded into bespoke morphologies, which limits their portability across platforms.

\begin{figure}
    \centering
    \includegraphics[trim={210 110 210 110},clip,width=0.87\linewidth]{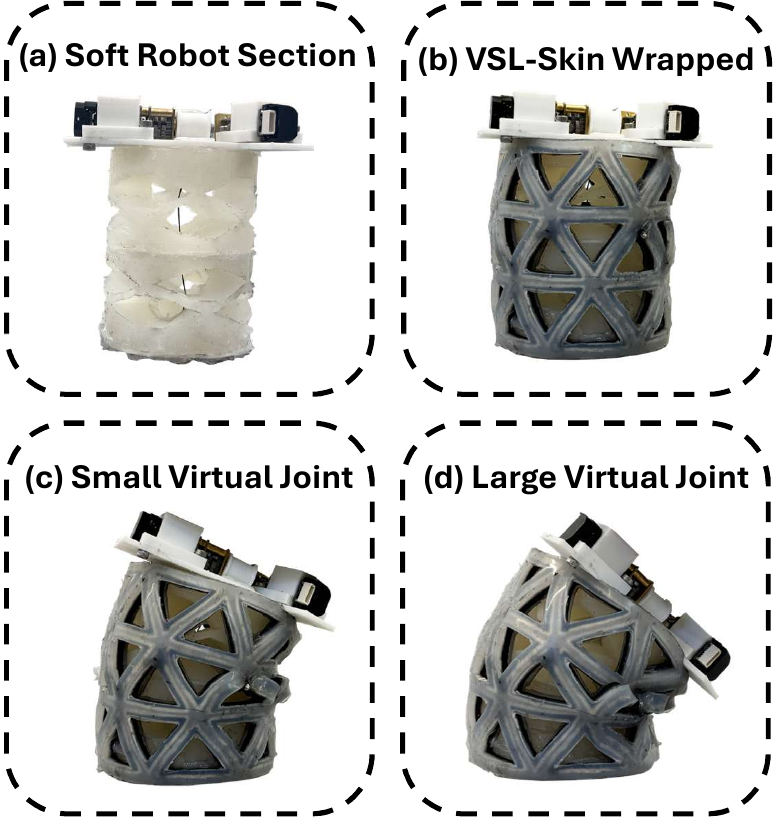}
    \caption{VSL-SKin allows dynamically-controllable morphologies through a modular design.}
    \label{fig:graphical_abstract}
\end{figure}

We present the Variable Stiffness Lattice Skin (VSL-Skin), a platform-agnostic solution enabling autonomous, spatially programmable fine-scale morphological adaptation. The proposed framework comprises a conformal triangular lattice comprising individually-controllable voxels. Each 18mm triangular voxel houses an LMPA structure thermally coupled to an embedded heater, enabling independent switching between rigid and compliant states via localized phase change. By selectively activating subsets of voxels, the skin defines centimeter-scale ``stiffness pixels'' that form virtual joints, impose curvature fields, and realize anisotropic responses (axial, bending, shear, torsion). Individual voxels are powered through a conventional harness (e.g., grouped branches with local switching), avoiding specialized bus topologies while maintaining per-voxel control. Tiling multiple sheets scales coverage without altering the per-voxel actuation principle.

The resulting skin provides: (i) \emph{spatial programmability} at centimeter scale, enabling placement of virtual hinges and compliant corridors along arbitrary geodesics on a host structure; (ii) \emph{multi-axis stiffness synthesis}, achieved by composing activation patterns that preferentially alter axial stiffness, bending stiffness about specified axes, in-plane shear, or torsional rigidity, with nearly two orders of magnitude modulation demonstrated across modes; (iii) \emph{dynamic reconfiguration}, where joint locations and stiffness gradients can be updated on demand without mechanical reassembly; and (iv) \emph{platform decoupling}, whereby a single skin can be trimmed, wrapped, or tiled while retaining full addressability under standard powering. These capabilities are quantified in Section~IV via mode-specific stiffness ratios, virtual-joint formation and pose-hold demonstrations, duty-cycle studies of thermal actuation (single-voxel heat/cool times on the order of tens of seconds in our prototypes), and trimming/integration experiments that preserve addressing and function.

Unlike variable-stiffness mechanisms embedded in custom morphologies, VSL-Skin externalizes stiffness modulation into a modular, reconfigurable layer that can be retrofit onto manipulators, wearables, and grippers. Localizing thermal mass to each voxel reduces unnecessary heat spread relative to bulk phase-change designs and improves switchability; the silicone lattice provides thermal isolation and mechanical compliance in the off state. The standard harness simplifies assembly and maintenance, supports fault isolation through straightforward continuity checks, and accommodates edge crops via re-termination. Because actuation is per-voxel, virtual joints degrade gracefully under single-voxel failure and can be rerouted by updating activation patterns. In practice, re-melting followed by re-solidification restores the mechanical state after overload, providing a thermal reset path that mitigates apparent “joint fatigue” by healing plastically deformed alloy slugs.

Overall, our contributions are: 
\begin{itemize}
    \item \textbf{VSL-Skin}, a robotic skin enabling individually addressable voxel-level stiffness control with precise modulation across axial, bending, shear, and torsional modes.
    \item A validated design framework combining closed-form beam–lattice models and finite element analysis for predictive stiffness synthesis and virtual joint placement.
    \item We demonstrate modulation across morphology under standard powering that maintains functionality after trimming, enabling platform-agnostic deployment and rapid prototyping.
    \item We present extensive experimental validation of programmable joint behaviors, including virtual hinges, shear compliance, and torsional articulation, via selective voxel activation patterns.
\end{itemize}

The remainder of the paper is organized as follows: Section~II reviews related work; Section~III details design, fabrication, and modeling; Section~IV reports experimental results and performance metrics; the conclusion is presented in Section~V.

\section{RELATED WORK}
\paragraph*{Granular jamming}
Granular jamming stiffens a compliant particle-filled bladder by increasing confinement, raising inter-particle friction, and establishing force-chain networks \cite{manti_2016_stiffening}. It underpins universal grippers, joint supports, and continuum/wearable robots that need shape retention \cite{brown_2010_universal,hauser_2017_jammjoint,liu_2021_a}. As a skin modality, the granular layers laminate onto the host structures to modulate the global dynamics and payload without requiring redesign of the body \cite{booth_2018_omniskins,shah_2020_jamming,bruder_2023_increasing}. However, granular systems require pumps and plumbing, exhibit hysteresis due to particle rearrangement, and offer patch-level rather than voxel-level addressability, which limits the precision of virtual joints on curved or trimmed substrates.

\paragraph*{Layer jamming}
Layer jamming raises bending and shear stiffness by compressing or clamping stacked sheets so that friction couples their motion. This produces a large reversible stiffness change with a low profile \cite{kim_2013_a,mitsuda_2017_variablestiffness,narang_2018_mechanically}. It is established in surgical tools, manipulators, and wearable supports actuated by pneumatic compression or electroadhesion \cite{kim_2013_a,wang_2019_electrostatic,choi_2019_soft}. Skin-specific embodiments retrofit stiffness control to objects and garments with minimal geometric changes \cite{ramachandran_2018_allfabric,booth_2018_omniskins,shah_2023_robotic,kwon_2022_selectively}. Limitations include high-voltage or vacuum infrastructure, frictional wear, and coarse granularity (patch scale), which restricts the location of torsional hinges and sharp joints.

\paragraph*{Material/phase-change approaches}
Material-centric methods change intrinsic modulus/yield behavior, melting /solidifying LMPA, activating shape memory materials, or exploiting reversible plasticity in architected lattices, to deliver compact and integration-friendly stiffness control \cite{cianchetti_2014_bioinspired,hwang_2022_shape}. LMPA systems, in particular, enable strong load transfer and rapid reconfiguration in continuum tools and manipulators (e.g., catheters with radial thermal gradients, arms driven by LMPA) and can be embedded in skins or sheaths with distributed heaters, including thermoelectric hybrids and LMPA \cite{lussi_2021_a,wang_2021_flexible,mccabe_2024_combining,song_2025_a,shamsaalharthy_2024_variable, manti_2016_stiffening}. However, LMPA approaches are often harder to control, lack voxel-level control equally, and are complex to integrate.

\section{THERMO–MECHANICAL MODELS}
This section formalizes the design logic of the VSL-skin that links geometry, kinematics, mechanics, and thermal actuation. Under a fixed-mass and manufacturability constraint, we adopt a stretch-dominated triangular voxel lattice and map its helical layout on a cylinder to an unwrapped sheet to quantify stroke and areal density, exposing the trade between voxel resolution \(N_{\theta}\) and compression ratio \(C\). We derive first-order stiffness and strength scalings that identify the buckling–yield transition as a function of ligament thickness \(t_f\) and span \(S_0\), and use these relations to set safe load and deflection targets. We then analyze heater geometry and voxel size to bound melt/cool time constants governing cycle rate, and show how patterned melting forms virtual joints with prescribed rotational stiffness. Finally, we report stiffness per unwrapped area to compare designs fairly across resolutions. Overall, the proposed framework parametrizes the performance and capability of the VSL-skin, enabling easy scaling and optimization.
\subsection{Geometry Selection}
\begin{figure}[t]
  \centering
  \includegraphics[width=0.8\linewidth]{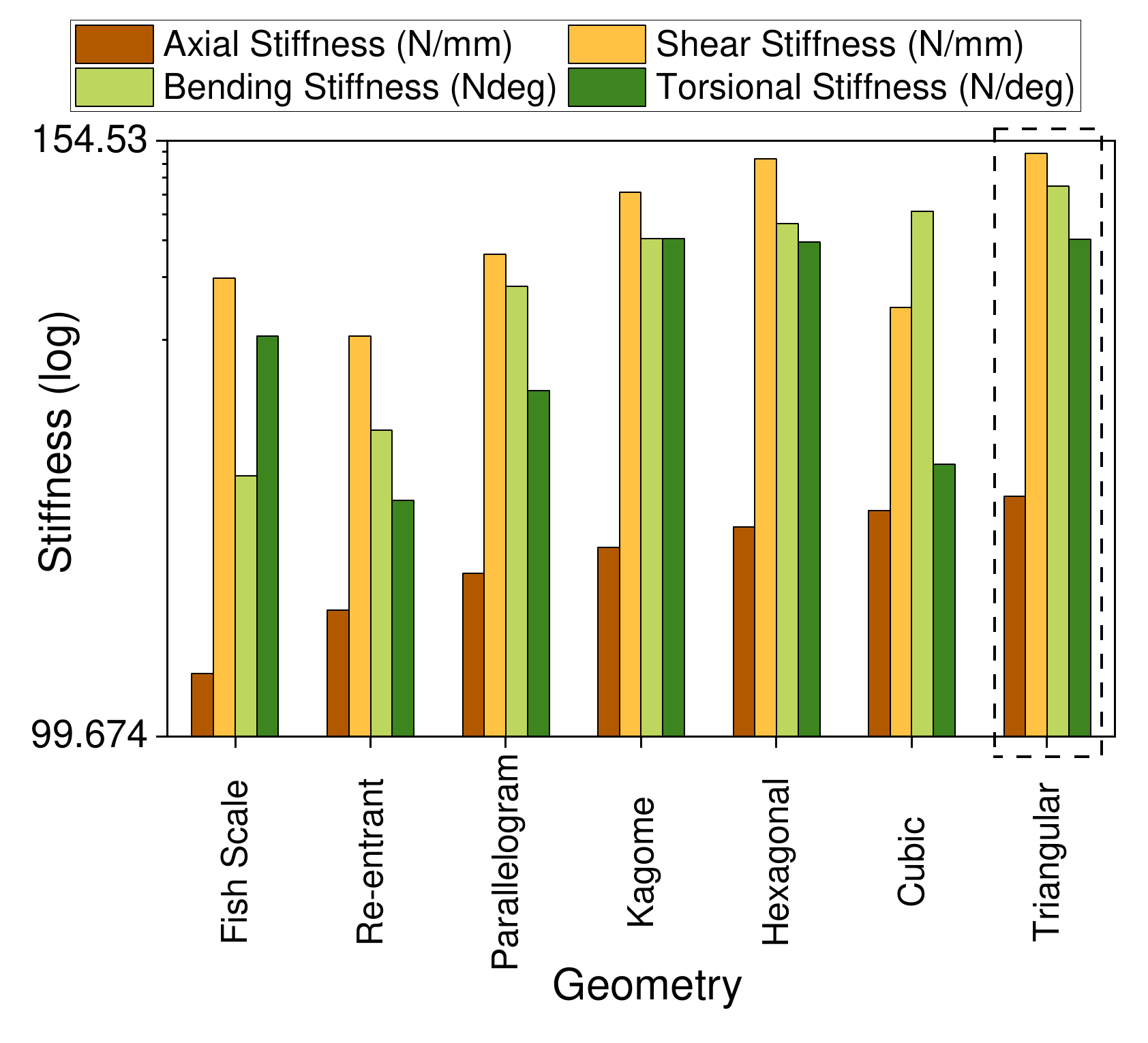}
  \caption{Normalized stiffness comparison of candidate lattices under fixed-mass constraints.}\vspace{-1.5em}
  \label{fig:GeometricJust}
\end{figure}

The goal for the VSL skin is to deliver highly programmable stiffness across axial, shear, bending, and torsional modes while preserving manufacturability under a fixed mass budget. Geometry selection is crucial because lattice topology governs specific stiffness, dominant failure mechanisms, and achievable stroke at a given areal density. We compared bands formed from seven planar lattices (cubic, fish-scale, hexagonal, Kagome, parallelogram, re-entrant, triangular) were compared on a 30\,mm cylinder with equalized mass (30\,g). The triangular voxel lattice achieved the highest normalized axial, shear, bending, and torsional stiffness (Fig.~\ref{fig:GeometricJust}) because it shortens unsupported spans and increases parallel load paths per area. The triangular voxel is adopted for subsequent design.

\subsection{Voxel Kinematics}
Due to the geometry of the triangular voxel, a geometric description of the VSL-SKin is required for size selection. To establish stroke bounds at a fixed areal density, we map the helical voxel layout to an unwrapped sheet and define the kinematic limits. A helical band of \(m\) turns on a cylinder of radius \(R\) with \(N_\theta\) voxels per turn defines a circumferential voxel edge length \(S_0\) equal to the azimuthal arc per voxel (total circumference divided by \(N_\theta\)). Fabrication reduces this to \(S_L<S_0\). With \(N_z\) stacked layers and interlayer stand-off \(h_0\),
\[
H\approx \frac{\sqrt{3}}{2}\,S_0 N_z + (N_z-1)\,h_0,\qquad
\Delta_{\max}\lesssim \frac{\sqrt{3}}{2}\,S_L N_z,
\]
neglecting residual thickness and contact. The compression ratio is approximately \(\mathcal{C}\approx S_L/S_0\) when \(h_0 \ll (\sqrt{3}/2)\,S_0\), revealing the central trade: increasing azimuthal resolution (larger \(N_\theta\)) shortens \(S_0\) and raises stiffness but reduces \(\mathcal{C}\). Also, increasing \(N_z\) scales height and stroke together while improving fabrication, increases \(S_L\), and recovers stroke at a given density.

\subsection{Strength and Stiffness}
Structural performance in the VSL-Skin framework sets allowable load, deflection, and governing failure modes at given areal densities, critical for operation reliability. Field’s–metal ligaments have thickness \(t_f=\phi_f t_{\mathrm{sheet}}\) within a silicone sheet of thickness \(t_{\mathrm{sheet}}\). The ligament width is \(\alpha S_0\), and the number of engaged paths scales as \(\sim N_\theta N_z\) (modulo orientation/engagement). Axial yield capacity grows with metallic cross‐section and path count, \(F_y \propto \sigma_y(T,\dot\varepsilon)\,\alpha S_0 t_f\, (N_\theta N_z)\). Treating a ligament as a column of span \(L\approx S_0\) and second moment \(I\sim \alpha S_0 t_f^{3}/12\) gives the Euler scaling \(F_{cr}\sim E\alpha t_f^{3}/S_0\). Thin, long ligaments are buckling controlled, thicker, shorter ligaments are yield controlled, with the transition primarily set by \(t_f\) and \(S_0\) (crossover when \(F_y\approx F_{cr}\)). For small strains, a ligament contributes bending stiffness \(k_b\sim t_f^{3}/S_0^{2}\) and axial stiffness \(k_s\sim t_f\). Summed over engaged paths, the effective thickness–length exponent is typically near \(2\)–\(3\) when bending and stretching both contribute. These relations map design knobs to performance: increasing \(t_f\), shortening \(S_0\), and raising the number of engaged paths (\(N_\theta N_z\)) increases safe load and stiffness at fixed mass.

\subsection{Thermal Actuation: Switching Time and Cycle Rate}
In addition to the structural performance, the activation and deactivation of each voxel will also need to be considered when designing. For the VSL skin, the cycle rate is bounded by the melt and cool times set by voxel geometry and heater layout. A perimeter heater of width \(\omega\) routed along the triangle edge length \(3S_0\) has resistance \(R_h \approx \kappa R_s (3S_0/\omega)\), where \(R_s\) is sheet resistance and \(\kappa\!\ge\!1\) accounts for meanders. For fixed thickness and fill fractions, the melt/fuse energy scales with planform area, \(Q_{\mathrm{melt}}\propto S_0^2\). With drive voltage \(V\) and total resistance \(R_h+R_{\mathrm{ser}}\),
\[
\tau_{\mathrm{melt}} \sim \frac{Q_{\mathrm{melt}}(R_h+R_{\mathrm{ser}})}{\eta V^2}
\ \propto\ \frac{S_0^3}{\omega V^2}
\quad \text{when } R_{\mathrm{ser}}\ll R_h,
\]
so smaller voxels and wider traces shorten actuation time, whereas larger voxels slow response. Cooling follows a lumped model with \(\tau_{\mathrm{cool}} = C_{\mathrm{th}}/G_{\mathrm{th}}\), where \(C_{\mathrm{th}}\) is set by voxel mass and \(G_{\mathrm{th}}\) by conduction to the substrate and ambient. Increasing \(G_{\mathrm{th}}\) or reducing mass decreases cycle time but can raise parasitic loss during heating.

\subsection{Resolution and Stiffness}
Additionally, the number of voxels per row (resolution) sets addressability and mediates the stiffness–stroke trade. Increasing \(N_\theta\) shortens \(S_0\), boosting stiffness per area superlinearly (per-path stiffness and path count both rise), while the compression ratio decreases as \(\mathcal{C}\approx S_L/S_0\). To compare designs when height and coverage vary, we report stiffness per unwrapped sheet area, \(k_{\mathrm{area}}=k/A_{\mathrm{sheet}}\) with \(A_{\mathrm{sheet}}=(2\pi m R)H\), which isolates topology/resolution effects at fixed areal density. This enables precise stroke declaration for mission planning accompanied by precise stiffness prediction.

\section{FABRICATION AND CALIBRATION}
\label{subsec:fabrication_calibration}
\begin{figure}[t]
    \centering
    \includegraphics[trim={0 150 0 170},clip,width=\linewidth]{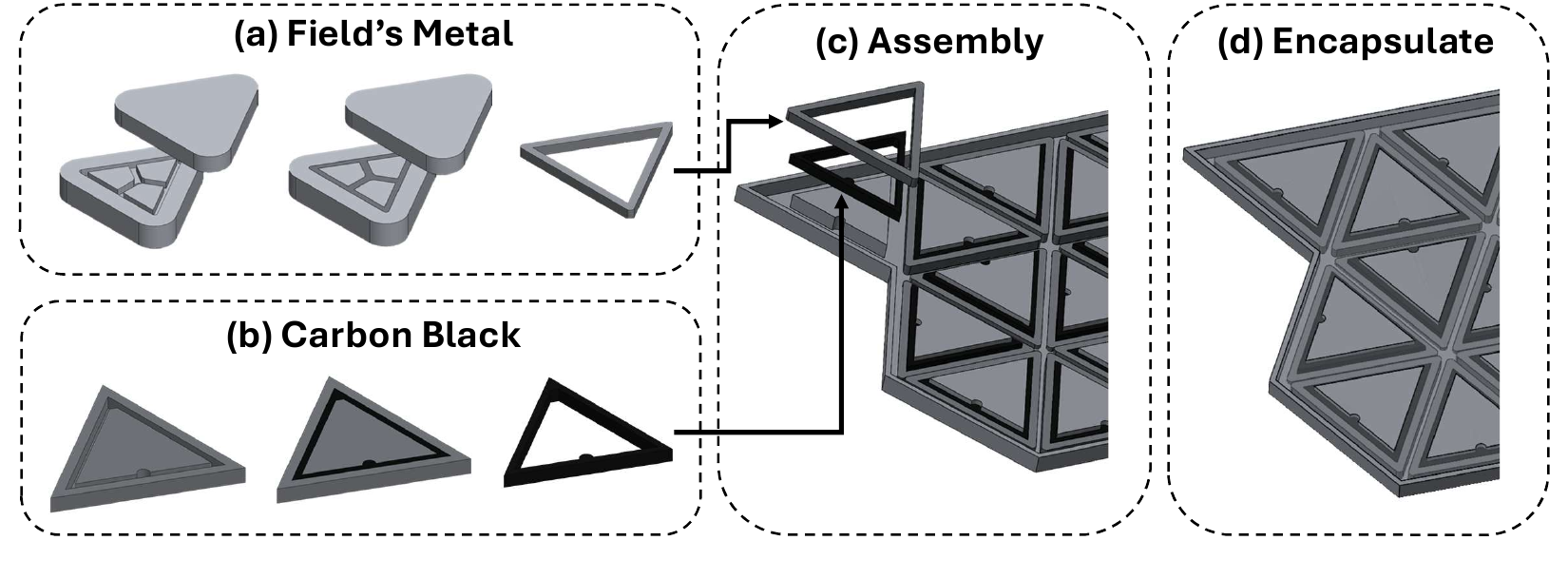}
    \caption{(a) Field's metal insert molding. (b) Carbon black heating element molding. (c) Assemble using the guide. (d) Encapsulate and seal components using Dragon Skin.} \vspace{-1.0em}
    \label{fig:fab}
\end{figure}

The VSL-Skin framework employs materials optimized for thermal switching. We use Carbon black (Super P) and heater matrix silicone (Ecoflex 30) for the resistive heater. For the body lattice, we use structural/encapsulation silicone (Dragon Skin 30). The Field’s metal (\(T_m\!\approx\!62^{\circ}\)C) is used for the liquid metal interface with embedded copper electrodes. Voxel fabrication begins from 3-D printed positive molds of the Field's metal structure. A silicone negative is cast, equipped with an injection port and vent channels to purge trapped air. The negative and alloy are held slightly above the melting temperature to ensure wetting. Then, the Field's metal is injected, and the filled mold is water-cooled for 2 minutes before demolding.

In parallel, the resistive heater is fabricated by incrementally mixing sieved and dried Carbon black into silicone part A to the target loading of 11 wt\%. The mixture is shear-processed to break agglomerates and vacuum-degassed to remove micro bubbles. Then, part B of the silicone is added, and the composite is pressed into the mold. The resistor is cured for 4 hours in a \SI{60}{\celsius} oven before demolding.

The components are then aligned onto a guide, and the entire structure is encapsulated to the target sheet thickness so the metal and the conductors are fully sealed. The structure is again cured for 4 hours in a \SI{60}{\celsius} oven before being released from the mold. Then, using a guide, the copper electrodes are inserted into the carbon black, ensuring proper contact geometry and contact pressure.

We then calibrate each voxel to ensure uniform thermal response across all voxels to account for fabrication inconsistency. The calibration procedure is presented in Algorithm~\ref{alg:voxel-cal}. The learned calibrations are stored to allow downstream autonomous control algorithms to scale trigger commands and time adequately. This adaptive approach overcomes material and fabrication variabilities while ensuring reliable performance across the grid.

\begin{algorithm}
\caption{Per-voxel thermal calibration}
\label{alg:voxel-cal}
\DontPrintSemicolon
\SetKwInput{KwInput}{Input}
\SetKwInput{KwOutput}{Output}
\KwInput{voxel set $\mathcal{V}$; duty grid $D$; $f_{\mathrm{PWM}}$; $f_s$; slope threshold $\epsilon$; max dwell $t_{\max}$}
\KwOutput{$\forall i\!\in\!\mathcal{V}$: $R_h[i],\,R_{\mathrm{tot}}[i],\,\tau_{\mathrm{th}}[i],\,d_i^\star(\cdot)$}
\For{$i \in \mathcal{V}$}{
  \tcp{Electrical}
  apply low-duty PWM; measure $(V,I)$; set $R_{\mathrm{tot}}[i]\leftarrow \operatorname{median}(V/I)$; estimate $R_h[i]$.\;
  \tcp{Duty sweep}
  \For{$d \in D$}{
    apply $\mathrm{PWM}(i,d,f_{\mathrm{PWM}})$; wait until $|\mathrm{d}T/\mathrm{d}t|<\epsilon$ or $t=t_{\max}$;\;
    record $\bar T_i(d)$ over a tail window.\;
  }
  fit an increasing map $T_i(d)$; define $d_i^\star(T)\leftarrow T_i^{-1}(T)$.\;
  \tcp{Step identification}
  choose small step $(d_0\!\to\! d_1)$; record $T_i(t)$ at $f_s$; fit first-order model; set $\tau_{\mathrm{th}}[i]$.\;
  store $\{R_h[i],R_{\mathrm{tot}}[i],\tau_{\mathrm{th}}[i],d_i^\star(\cdot)\}$.\;
}
\end{algorithm}

\section{EXPERIMENTS AND RESULTS}
We validate the VSL-Skin framework through systematic simulation and characterization of its morphological adaptation capabilities. Our validation demonstrates the framework's ability to achieve precise, voxel-level control over stiffness fields while maintaining platform-agnostic deployment characteristics. These results establish the first adaptive system, to our knowledge, capable of centimeter-scale stiffness programming across multiple mechanical modes.

\begin{figure}[b]
    \centering
    \includegraphics[trim={20 10 20 10},clip,width=.89\linewidth]{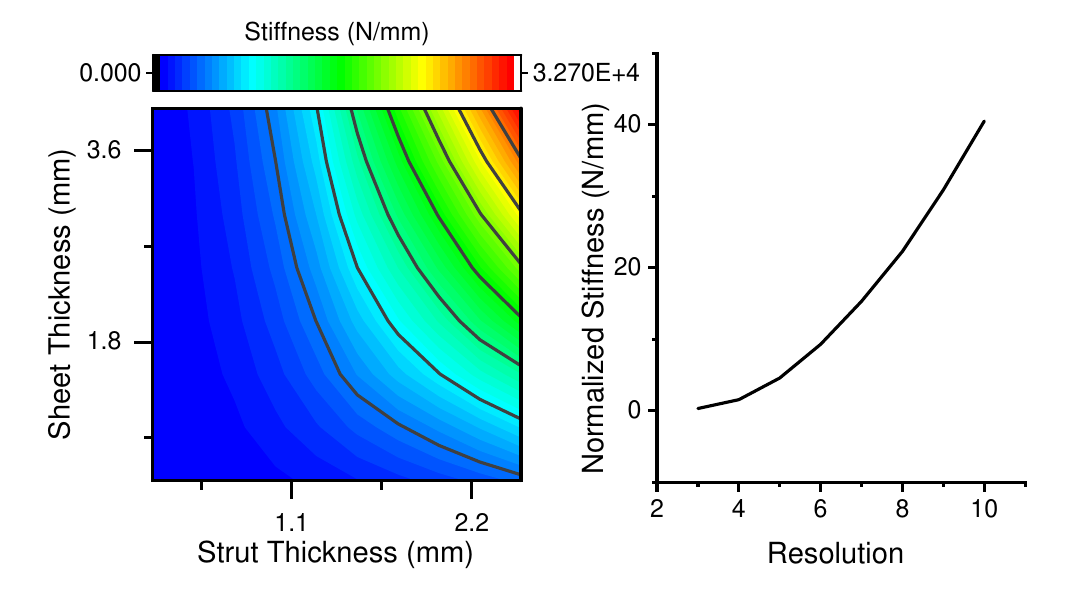}
    \caption{Parametric studies of voxel-level design. (Left) Variation of thickness; (Right) variation of angular resolution.}
    \label{fig:geomParam}
\end{figure}

\subsection{Design Parameter Control and Scalability}
As established in our thermo-mechanical models, the two main adjustable parameters are strut thickness and overall skin thickness. We systematically varied the strut thickness \(t_f\) and skin thickness \(t_{sheet}\) and present their relationship in Fig~\ref{fig:geomParam}. When conducting the simulation, we modeled the shape without encapsulated silicone, as the main support is derived from the LMPA. Deactivated cells are modeled with the material Dragon Skin. 

For strut thickness, we observed that rigidity scales \(k \propto t_f^{2.52}\) (shown in Fig.~\ref{fig:geomParam}). This departs from the ideal Euler-Bernoulli scaling of \(t_f^{3}\) due to mixed-mode response combining bending, membrane stretching, and transverse shear from wrapped curvature. Skin thickness contributes approximately linearly with \(k \propto t_{\mathrm{sheet}}^{1.11}\), consistent with a membrane constraint that suppresses local rotations and redistributes load across neighboring voxels. From a stiffness-maximizing perspective, strut thickness should be the primary parameter to optimize. However, increasing strut thickness will reduce max compression. Thus, a balance between strut thickness and skin thickness can be identified for each specific task through the iso-stiffness curve presented.

Another parameter to consider is the skin stiffness resolution. Our simulation analysis reveals that normalized stiffness scales almost quadratically with the number of voxels per turn as \(k_{\mathrm{norm}} \propto N_\theta^{2}\). Another tradeoff can be identified, with \(S_0 \propto 1/N_\theta\), the melt time follows \(\tau_{\mathrm{melt}} \propto N_\theta^{-3}\) and the compression ratio decreases as \(\mathcal{C}\!\approx\!S_L/S_0\).Resolution follows a similar design law as strut thickness. Thus, another balance can be struck, allowing for dynamic optimization for different tasks. For example, if the task calls for maximized strength, resolution and strut thickness should be maximized.

\begin{figure}[b]
    \centering
    \includegraphics[width=0.8\linewidth]{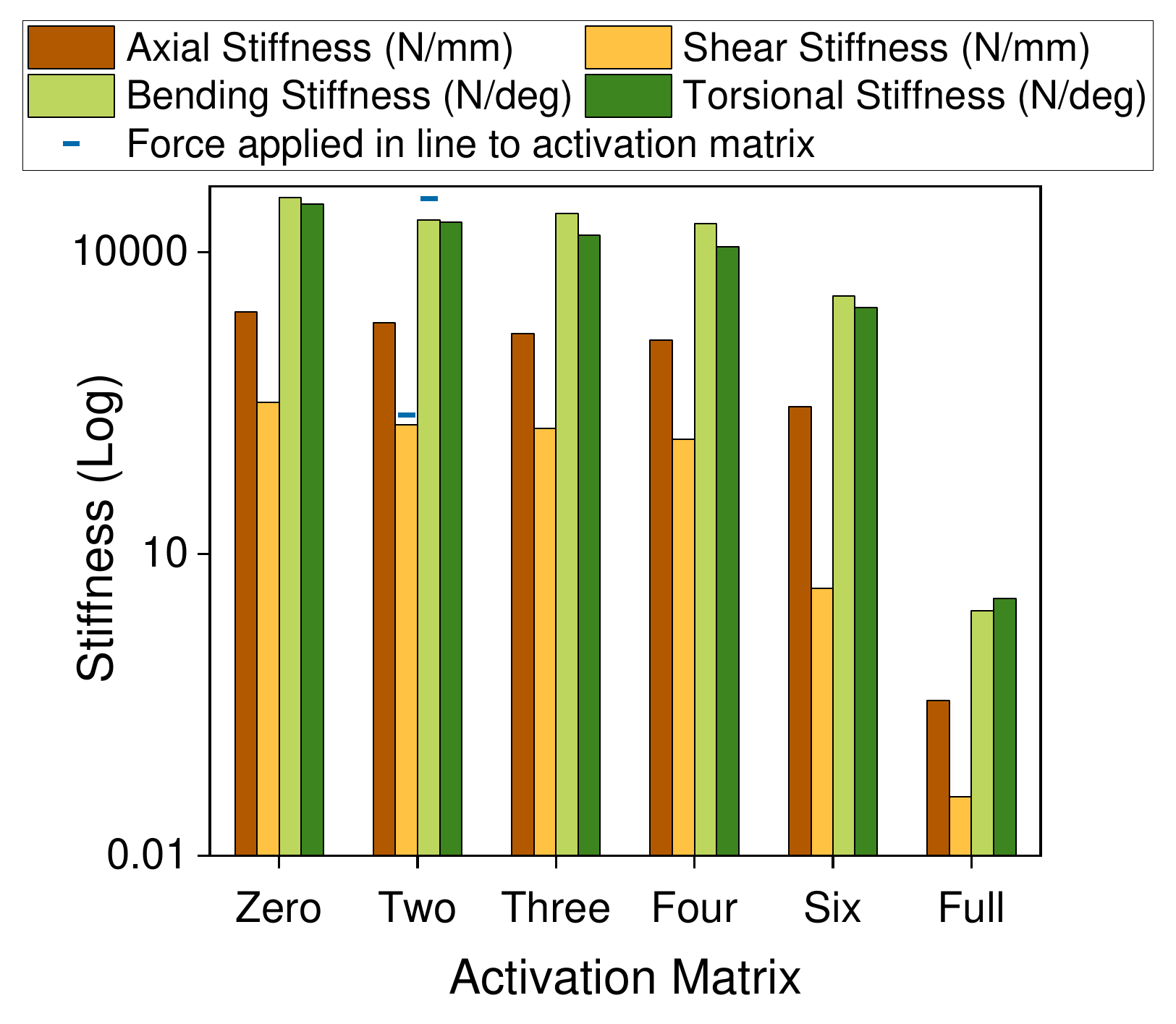}
    \caption{Chart (log scale) showing the programmable stiffness range across axial, shear, bending, and torsional modes.}
    \label{fig:stiffnessResults}
\end{figure}

VSL-Skin uses a voxel-indexed control abstraction in a 2D lattice coordinate frame. Electrically, this abstraction can be implemented as direct per-voxel switching. Scaling is primarily constrained by peak electrical power during concurrent heating and by heat rejection during cooldown; as a result, large arrays benefit from activation scheduling that staggers voxel heating to respect supply limits. The skin can be tiled to increase coverage without changing the per-voxel actuation principle. Trimming preserves function by re-terminating exposed conductors and updating the voxel map used by the controller, all made possible by the modular voxel design.

\subsection{Stiffness Modulation Across Modes}
The individually addressable voxel enables variable stiffness modulation. We demonstrated VSL-Skin's stiffness control spanning nearly two orders of magnitude across axial, shear, bending, and torsional modes. Fig.~\ref{fig:stiffnessResults} presents the complete stiffness envelope achievable through the VSL-Skin framework. Across all mechanical modes, monotonic modulation from fully solid to fully melted states spans the regimes required for autonomous task-level specification. The discrete activation sets (Zero, Two, Three, Four, Six, Twelve) demonstrate stepwise control over stiffness without requiring reconfiguration of host structures. This enables further precise and predictable control.

Further control can be gained with specific activation patterns. In the “Two” pattern, in-line loading increases shear from \(194\) to \(240~\mathrm{N/mm}\) (\(+24\%\)) and bending stiffness from \(20.9\times10^{3}\) to \(33.9\times10^{3}~\mathrm{N/deg}\) (\(+62\%\)), effectively biasing deformation along the programmed axis. This capability establishes a predictable mapping from activation patterns to torque-angle targets and local compliance fields, validating our approach to virtual hinges and anisotropic joints.

\subsection{Virtual Joint Generation and Morphological Control}
\begin{figure}[!t]
    \centering
    \includegraphics[trim={110 20 120 0},clip, width=0.98\linewidth]{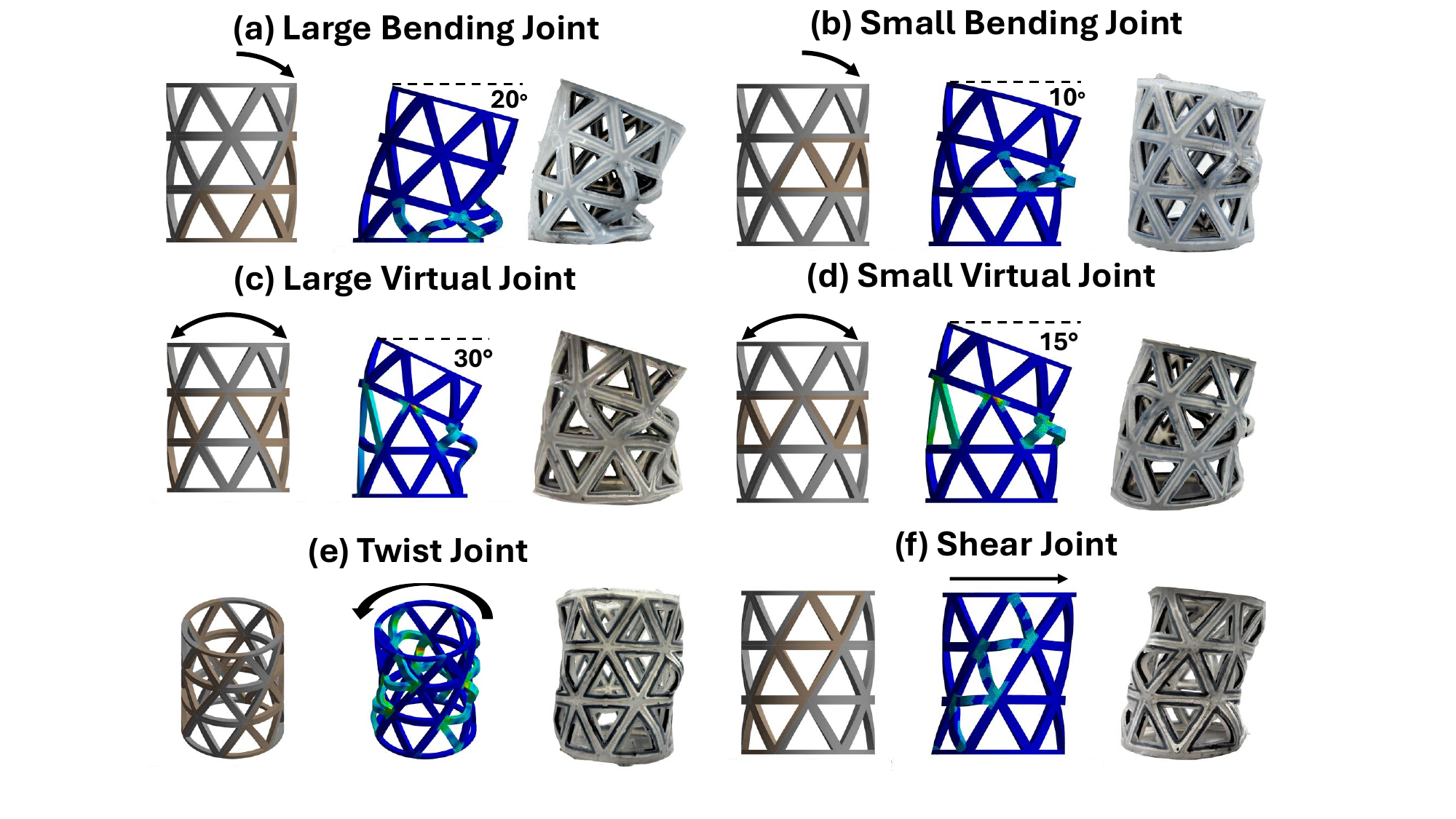}
    \caption{Six programmable joint configurations (a–f). Each panel shows \textbf{left}: activation matrix (brown = activated, gray = deactivated), \textbf{middle}: simulated elastic strain field, and \textbf{right}: experimental deformed shape. Strain localizes to the activated voxels, enabling placement-specific compliance.}\vspace{-1.5em}
    \label{fig:jointsGrid}
\end{figure}

To demonstrate the ability to form virtual joints, we evaluate six canonical joint patterns, including bending, virtual hinges, localized torsion, and lateral shear, to establish the framework's morphological adaptation capabilities. Fig.~\ref{fig:jointsGrid} shows the voxel-level placement and mode selection across these configurations. In all cases, strain remains confined to the activated region while neighbors stay stiff, matching the virtual-joint model.

\paragraph*{Bending joints (\emph{unilateral})} Configurations (Fig.~\ref{fig:jointsGrid} (a,b)) achieve hinge-like rotation with tunable magnitude through activation-band width and column count. Widening melted bands increases rotation while reducing rotational stiffness. The localization effect keeps non-activated columns stretch-dominated, enabling multiple bends on the same sheet with minimal cross-talk for precise shaping of curvature fields around task contacts. The joint spans only one side, practical when motion must be biased away from obstacles.

\paragraph*{Virtual joints (\emph{bilateral})} Configurations (Fig.~\ref{fig:jointsGrid} (c,d)) create a concentrated rotation zone (virtual hinge) by activating a symmetrical pattern. The larger pattern achieves (\(30^\circ\)) rotation with lower stiffness than the smaller (\(15^\circ\)) pattern, consistent with the series addition of softened segments and the thickness–span scalings established earlier. Curvature concentrates precisely where specified, while axial support is retained, enabling directional steering on bodies with limited mounting areas. In-line bending is also geometrically disfavored, enabling precise pose prediction and control. The resulting workspace is two-sided and symmetric about the zero band, which is advantageous for steering and pose regulation that require reversible curvature.

\paragraph*{Twist} Configurations
(Fig.~\ref{fig:jointsGrid} (e)) achieves localized torsion by staggering activation circumferentially, biasing the lattice so rotation about the surface normal dominates with minimal axial extension. The simulated strain gradients center on the activated zone and align with the observed twist, providing torsional articulation without added couplers and complementing bending for out-of-plane orientation control. In addition, axial compression is discouraged, enabling fine control. The number of activated columns can be further constrained, enabling max twist angle modulation. The twist joint is useful when fine orientation control of an end-mounted sensor or tool is required, where the overall length and support must be preserved.

\paragraph*{Shear} Configurations
(Fig.~\ref{fig:jointsGrid} (f)) allows lateral translation while preserving axial rigidity by activating a narrow, oblique band. The skin is skewed toward the activation line as neighboring columns remain stiff, yielding high shear-to-axial selectivity for controllable slip at the skin–environment interface without sacrificing overall load support. The shear joint is applicable for tasks such as seating and centering an object in a gripper palm sideways while maintaining normal force, or biasing footpad traction to execute small lateral steps without changing upstream joint angles.

Collectively, these six patterns demonstrate precise placement, anisotropic mode shaping, and low cross-talk using only activation patterns—building blocks for richer compositions with the task-driven synthesis framework. These joints all show further control refinement, demonstrated by the ability to form large and small joints and control the maximum twist angle. Additional patterns are also possible through different activation matrices.

\begin{figure}[t]
    \centering
    \includegraphics[trim={0 100 0 100},clip, width=0.75\linewidth]{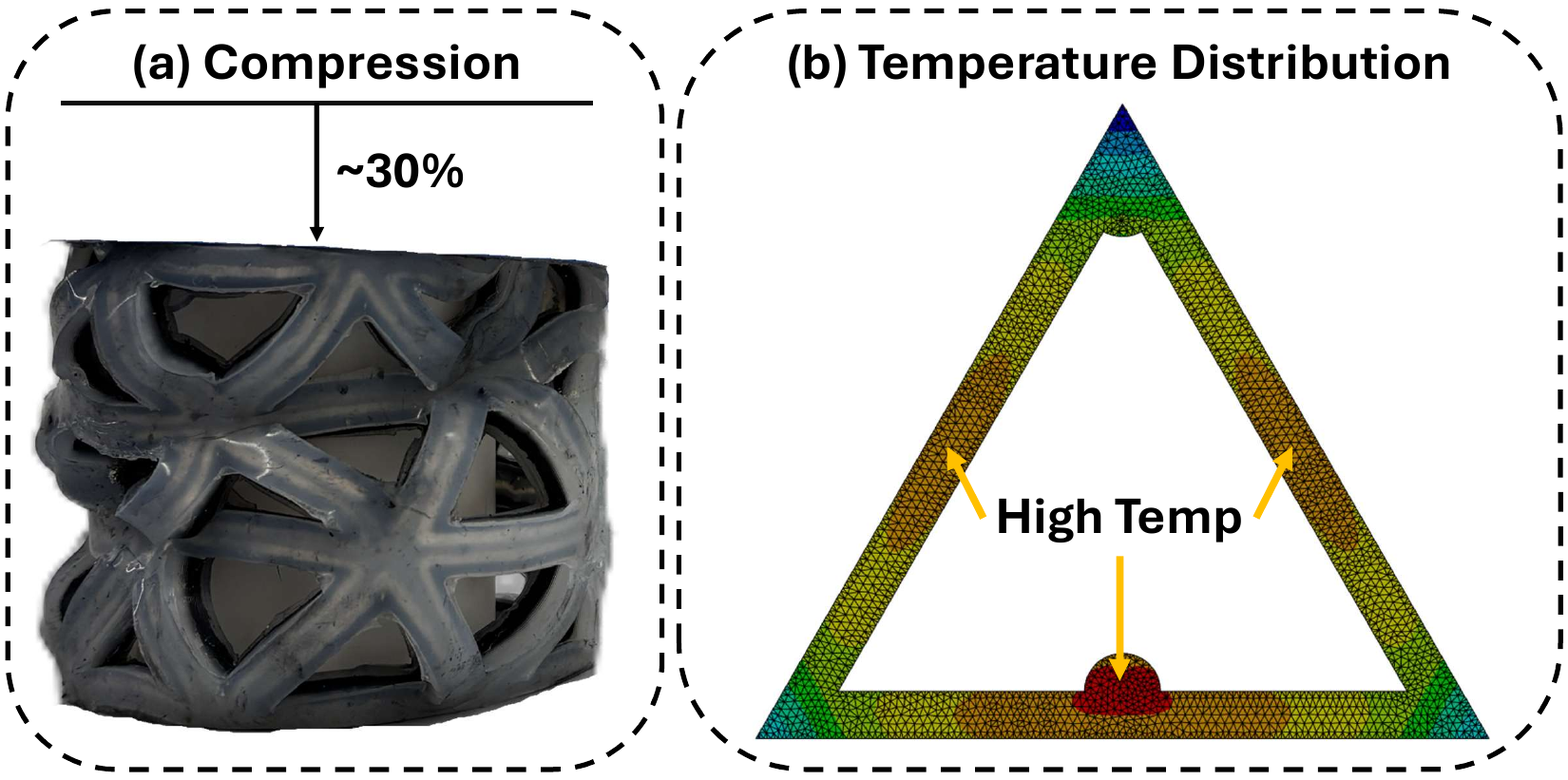}
    \caption{Programmable axial shortening and thermal Field. (a) Experimental contraction of the lattice sleeve under activation of longitudinal bands, achieving 30\% axial shortening. (b) Thermo-electric simulation of a voxel during resistive heating showing mid-span warming ahead of the vertices.}\vspace{-1.0em}
    \label{fig:compress_thermal}
\end{figure}

\begin{table}[!t]
\centering
\caption{Stiffness-modulation Benchmarking}
\label{tab:benchmark_minset}
\setlength{\tabcolsep}{2.5pt}
\scriptsize
\begin{tabular}{l l r l c c c}
\toprule
\textbf{Form} & \textbf{System} & \textbf{Mod.} & \textbf{Control} & \textbf{DL} & \textbf{LC} & \textbf{VJ} \\
\midrule
Phase-ch. &
Al-Harthy \emph{et al.}~\cite{AlHarthy2025PhaseChangeAlloys} &
${\sim}43\times$ &
Thermal &
\ding{55} & \ding{55} & \ding{51} \\
Lattice, &
\textbf{VSL-Skin*} &
${\sim}80\times$\textsuperscript{a} &
Joule (PWM) &
\ding{51} & \ding{51} & \ding{51} \\[-0.3ex]
phase-ch. & & & & & & \\
Layer-jam. &
Shah \emph{et al.}~\cite{Shah2021JammingSkins} &
${\sim}47\times$ &
Vacuum &
\ding{55} & \ding{55} & \ding{51} \\
Gran.-jam. &
Hauser \emph{et al.}~\cite{Hauser2017JammJoint} &
${>}7\times$ &
Vacuum &
\ding{55} & \ding{55} & \ding{55} \\
Fiber-jam. &
Yang \emph{et al.}~\cite{Yang2021TensileJamming} &
${>}20\times$\textsuperscript{b} &
Vacuum &
\ding{55} & \ding{55} & \ding{51} \\
\bottomrule
\end{tabular}
\vspace{1pt}

\raggedright
\scriptsize
Mod.\ = stiffness modulation; DL = damage localization; LC = local control;\\ VJ = virtual joints.
\textsuperscript{a}Axial; shear ${\sim}19\times$, bending ${\sim}38\times$.
\textsuperscript{b}Tensile; bending ${\sim}2\times$.\vspace{-1.0em}
\end{table}


\subsection{Morphological Adaptation and Thermal Management}
To our knowledge, no phase change stiffness modulation can achieve compression. Fig.~\ref{fig:compress_thermal}(a) demonstrates axial compression capabilities. Through simultaneous activation of all circumferential voxels, the sleeve enters a uniformly compliant axial state, enabling precise height control. Our current design achieves up to 30\% axial shortening. Re-solidification in the shortened pose restores high stiffness. Therefore, by controlling the number of activated rows, precise compression can be tuned for specific demands.

In addition to stiffness modulation, joint creation, and compression, thermal field analysis during resistive heating was conducted to demonstrate the nuances of the triangular heating component due to inherent uneven electrode placement. Fig.~\ref{fig:compress_thermal}(b) reveals that midsections of the triangular resistors heat first, followed by corners, arising from local Joule heating balances that produce nonuniform melt fronts. This behavior is addressed through our autonomous control software, which generates per-voxel PWM duty schedules to equalize melt fronts during group activations, ensuring uniform morphological adaptation and predictability.

In practice, voxels heat in approximately 30 seconds and cool in approximately 45 seconds. Because actuation operates per-voxel, energy is confined to the activated mass rather than the entire sheet. Unlike previous designs, simple spatial scheduling increases effective autonomous throughput within power and temperature budgets, establishing practical operational parameters for real-world deployment.

\subsection{Robustness and Self-Repair}
\begin{figure}[t]
    \centering
    \includegraphics[trim={25 80 25 80},clip,width=0.85\linewidth]{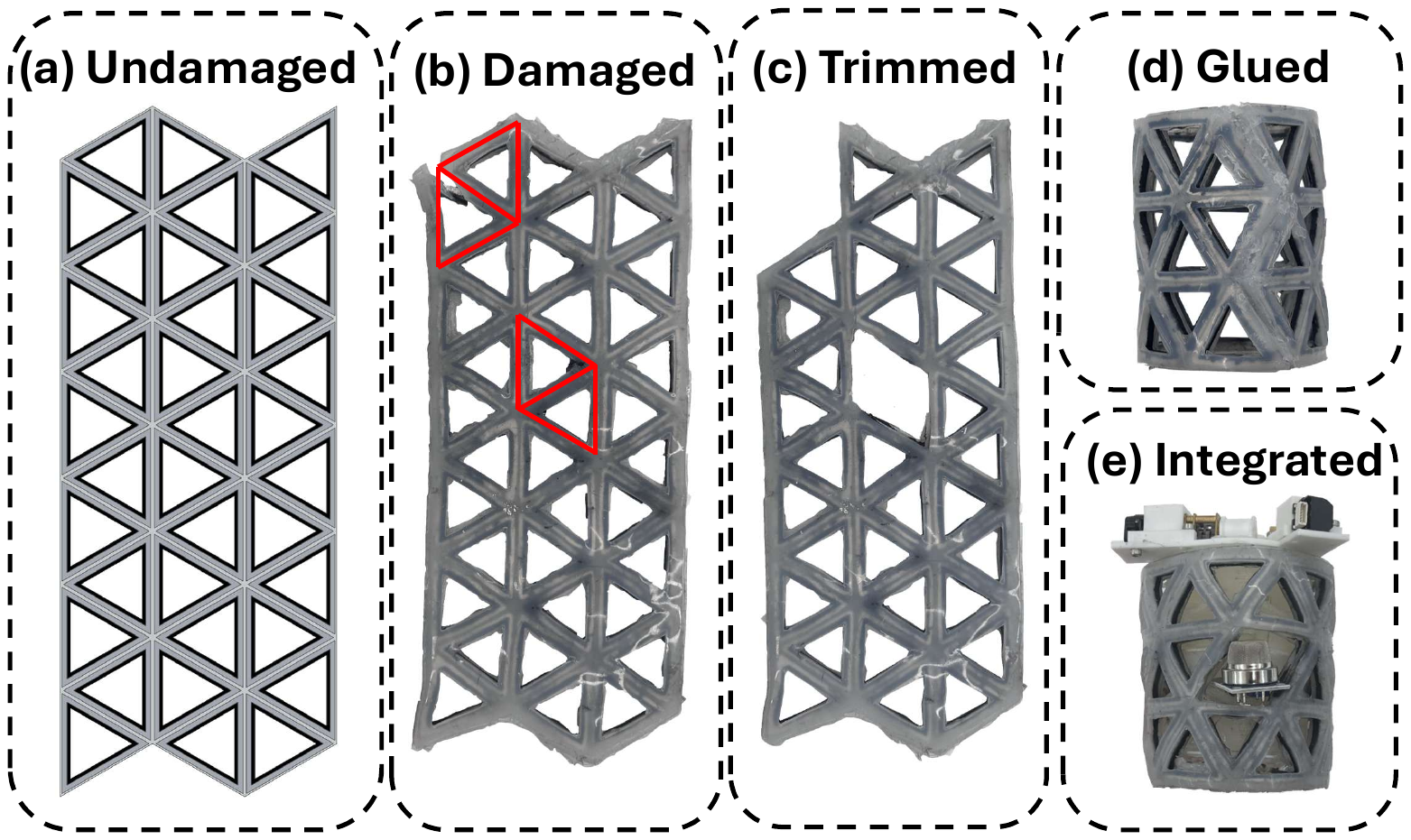}
    \caption{Damage–rework–integration of the VSL skin: (a) CAD; (b) deliberate four-voxel fracture on the fabricated skin; (c) trimmed aperture; (d) longitudinal adhesive seam; (e) installation on a robot arm with a through-axis sensor.}\vspace{-1.0em}
    \label{fig:trim}
\end{figure}

We validate the robustness and repairability of the VSL-Skin as demonstrated in Fig.~\ref{fig:trim}. Unlike previous designs, if the skins are damaged, the entire rehabilitation is not needed. The broken cells can be trimmed, and autonomous leak checks during repeated thermal cycling and bending showed no LMPA loss. The skin is then glued and integrated onto a standard robot arm and resumes normal function.

In addition to damage isolation, the skin supports in situ self-repair. After an overload, fractures can be formed within the Field's metal strut. These fractured voxels can be repaired through heat cycling. After one cycle, the voxel returns all functionality, enabling on-demand repair. This is highly useful for overload-tolerant manipulation. If the joints exceed the load limit, these joints can be reset on demand, maintaining availability and uptime, a significant improvement from traditional joints.

Furthermore, the principal fatigue failure in conventional joints is through cumulative material damage in load-bearing members. In VSL-Skin, reheating removes this failure accumulation completely. As a result, the system, in theory, can achieve longer service life and higher mission availability.

Due to this reusability and self-repairability, intentional sacrificial sections are possible, where localized tiles are tuned to yield to protect upstream structures and expensive components. By designing thinner LMPA struts in designated voxels near impact points, gripper edges, or cable exits, utilizing the proposed design laws, these sections can fail preferentially under load. After the event, the sacrificial voxels can be reheated to restore functionality, returning skin and main structure to service with minimal downtime. This capability has been impractical so far for soft robots, where elastomer skins tend to require recasting, and costly for traditional rigid robots that also demand part replacement. In contrast, VSL-Skin can provide predictable, repeatable sacrificial joints that are rapidly re-dployable.

\subsection{Modulation across Morphology and Platform Agnostic}
The row-column voxel architecture generalizes cleanly across different platforms and morphologies. The same control abstraction will work on a variety of different surfaces, even after trimming. Additional sheets can be connected or disconnected easily. Since addressability is per-voxel, common joint libraries can be shared, further easing integration.

Additionally, combinations of skin can be fitted together as composite envelopes, with sectional specialization. For example, one section can be tuned for high stiffness and pose-hold (high resolution and thick struts), and paired with sections for fast activation (low resolution and thin struts), creating robotic arms. The local thermal budget also ensures power consumption regulation, enabling mass scaling.

Through a simple scheduler, the VSL-Skin can be turned into a time-varying, multi-segmented arm with a combination of different skins through regulating and relocating multiple virtual joints dynamically, while preserving load paths and operating power envelope. By combining joints (location, type, magnitude, duration) into time-phased voxel commands, the surface can be represented through closed-loop kinematics. Since patterns are expressed in row-column frames, the same joint program ports across multiple different morphologies, enabling easy scaling and integration.

\subsection{Behchmarking}
Table~\ref{tab:benchmark_minset} compares VSL-Skin with representative variable-stiffness families using stiffness-modulation ratios. Here, $^*$ denotes the maximum modulation ratio among the listed systems, and “virtual joints” indicates the ability to create joint-like kinematic behavior through stiffness patterning (not necessarily a discrete mechanical hinge).

\section{CONCLUSION}

VSL-Skin fundamentally transforms morphological adaptation from passive compliance to active intelligence, establishing the framework for real-time, spatially programmable mechanical reconfiguration at voxel resolution. We established centimeter-scale stiffness programmability with nearly two orders of magnitude modulation across axial, shear, bending, and torsional modes while realizing compact joint sets and axial contraction on the same hardware platform. Our modeling-to-synthesis workflow links lattice geometry and key design parameters to target stiffness and stroke, producing pattern prescriptions that transfer across platforms without modification. The demonstrated autonomous per-voxel operation with 30-45 second thermal cycles and greater than 30\% axial shortening indicates readiness for manipulation and wearable applications. Moreover, the capability to realize sacrificial joints enables predictable failure management as localized voxels designed to fail preferentially under overload protect expensive components. We restore full functionality through simple reheating, transforming maintenance from reactive replacement to proactive thermal management. The platform-agnostic architecture with cut-to-fit integration preserves full addressability after trimming while enabling deployment across diverse robotic systems using standard harnesses, establishing immediate readiness for manipulation, wearable, and adaptive applications.

This work establishes morphological intelligence as a systematically engineerable capability, demonstrating that dynamic structural reconfiguration can be implemented at scale to advance adaptive system design across engineering domains. Future work will improve heat rejection, characterize long-term endurance, and integrate the joint library with task-level planners that automatically select design parameters and activation schedules for optimal autonomous adaptation. The established framework provides the foundation for next-generation adaptive robotic systems capable of real-time morphological reconfiguration to match task demands.
\vspace{-1.0em}

\bibliographystyle{ieeetr}
\bibliography{references}

@article{cianchetti_2014_soft,
  author = {Cianchetti, Matteo and Ranzani, Tommaso and Gerboni, Giada and Nanayakkara, Thrishantha and Althoefer, Kaspar and Dasgupta, Prokar and Menciassi, Arianna},
  month = {06},
  pages = {122-131},
  title = {Soft Robotics Technologies to Address Shortcomings in Today's Minimally Invasive Surgery: The STIFF-FLOP Approach},
  doi = {10.1089/soro.2014.0001},
  volume = {1},
  year = {2014},
  journal = {Soft Robotics}
}

@article{manti_2016_stiffening,
  author = {Manti, Mariangela and Cacucciolo, Vito and Cianchetti, Matteo},
  month = {09},
  pages = {93-106},
  title = {Stiffening in Soft Robotics: A Review of the State of the Art},
  doi = {10.1109/mra.2016.2582718},
  volume = {23},
  year = {2016},
  journal = {IEEE Robotics \& Automation Magazine}
}

@article{cianchetti_2014_bioinspired,
  author = {Cianchetti, Matteo and Licofonte, Alessia and Follador, Maurizio and Rogai, Francesco and Laschi, Cecilia},
  month = {07},
  pages = {226-244},
  title = {Bioinspired Soft Actuation System Using Shape Memory Alloys},
  doi = {10.3390/act3030226},
  volume = {3},
  year = {2014},
  journal = {Actuators}
}

@article{candia_2024_mutable,
  author = {Candia, Daniela and Michela Sugni and Bonasoro, Francesco and Wilkie, Iain C},
  month = {01},
  pages = {37-37},
  publisher = {Multidisciplinary Digital Publishing Institute},
  title = {Mutable Collagenous Tissue: A Concept Generator for Biomimetic Materials and Devices},
  doi = {10.3390/md22010037},
  urldate = {2024-05-06},
  volume = {22},
  year = {2024},
  journal = {Marine drugs}
}

@article{bruder_2023_increasing,
  author = {Bruder, Daniel and Graule, Moritz A. and Teeple, Clark B. and Wood, Robert J.},
  month = {08},
  publisher = {American Association for the Advancement of Science (AAAS)},
  title = {Increasing the payload capacity of soft robot arms by localized stiffening},
  doi = {10.1126/scirobotics.adf9001},
  volume = {8},
  year = {2023},
  journal = {Science Robotics}
}

@article{liu_2021_a,
  author = {Liu, Tianxin and Xia, Haisheng and Dae Young Lee and Amir Firouzeh and Park, Yong-Lae and Cho, Kyu-Jin},
  month = {07},
  pages = {8078-8085},
  publisher = {Institute of Electrical and Electronics Engineers},
  title = {A Positive Pressure Jamming Based Variable Stiffness Structure and its Application on Wearable Robots},
  doi = {10.1109/lra.2021.3097255},
  urldate = {2023-05-01},
  volume = {6},
  year = {2021},
  journal = {IEEE robotics and automation letters}
}

@article{kim_2013_a,
  author = {Kim, Yong-Jae and Cheng, Shanbao and Kim, Sangbae and Iagnemma, Karl},
  month = {08},
  pages = {1031-1042},
  title = {A Novel Layer Jamming Mechanism With Tunable Stiffness Capability for Minimally Invasive Surgery},
  doi = {10.1109/tro.2013.2256313},
  volume = {29},
  year = {2013},
  journal = {IEEE Transactions on Robotics}
}

@article{wang_2019_electrostatic,
  author = {Wang, Tao and Zhang, Jinhua and Li, Yue and Hong, Jun and Wang, Michael Yu},
  month = {04},
  pages = {424-433},
  title = {Electrostatic Layer Jamming Variable Stiffness for Soft Robotics},
  doi = {10.1109/tmech.2019.2893480},
  urldate = {2022-01-17},
  volume = {24},
  year = {2019},
  journal = {IEEE/ASME Transactions on Mechatronics}
}

@article{lussi_2021_a,
  author = {Lussi, Jonas and Mattmann, Michael and Semih Sevim and Grigis, Fabian and Carmela De Marco and Christophe Chautems and Pané, Salvador and Josep Puigmartí-Luis and Boehler, Quentin and Nelson, Bradley J},
  month = {07},
  pages = {2101290-2101290},
  publisher = {Wiley-Blackwell},
  title = {A Submillimeter Continuous Variable Stiffness Catheter for Compliance Control},
  doi = {10.1002/advs.202101290},
  urldate = {2023-08-23},
  volume = {8},
  year = {2021},
  journal = {Advanced Science}
}

@article{booth_2018_omniskins,
  author = {Booth, Joran W. and Shah, Dylan and Case, Jennifer C. and White, Edward L. and Yuen, Michelle C. and Cyr-Choiniere, Olivier and Kramer-Bottiglio, Rebecca},
  month = {09},
  title = {OmniSkins: Robotic skins that turn inanimate objects into multifunctional robots},
  doi = {10.1126/scirobotics.aat1853},
  url = {https://robotics.sciencemag.org/content/3/22/eaat1853},
  volume = {3},
  year = {2018},
  journal = {Science Robotics}
}

@article{kwon_2022_selectively,
  author = {Kwon, Junghan and Choi, Inrak and Park, Myungsun and Moon, Jeongin and Jeong, Bomin and Pathak, Prabhat and Ahn, Jooeun and Park, Yong‐Lae},
  month = {03},
  pages = {2101543},
  title = {Selectively Stiffening Garments Enabled by Cellular Composites},
  doi = {10.1002/admt.202101543},
  urldate = {2022-04-02},
  year = {2022},
  journal = {Advanced Materials Technologies}
}

@article{mitsuda_2017_variablestiffness,
  author = {Mitsuda, Takashi},
  month = {06},
  pages = {364-369},
  publisher = {IEEE},
  title = {Variable-stiffness Sheets Obtained using Fabric Jamming and their applications in force displays},
  doi = {10.1109/whc.2017.7989929},
  url = {https://ieeexplore.ieee.org/document/7989929},
  urldate = {2025-08-06},
  year = {2017},
  journal = {2017 IEEE World Haptics Conference (WHC)}
}

@article{zhang_2024_design,
  author = {Zhang, Jingyu and Fang, Qin and Liu, Lilu and Jin, Rui and Xiang, Pingyu and Xiong, Rong and Wang, Yue and Lu, Haojian},
  month = {01},
  pages = {1-11},
  publisher = {Institute of Electrical and Electronics Engineers},
  title = {Design and Stiffness Control of a Variable-Length Continuum Robot for Endoscopic Surgery},
  doi = {10.1109/tase.2024.3418092},
  url = {https://ieeexplore.ieee.org/document/10579861},
  urldate = {2025-08-06},
  year = {2024},
  journal = {IEEE Transactions on Automation Science and Engineering}
}

@article{wang_2021_flexible,
  author = {Wang, Haibo and Chen, Zhiwei and Zuo, Siyang},
  month = {06},
  title = {Flexible Manipulator with Low-Melting-Point Alloy Actuation and Variable Stiffness},
  doi = {10.1089/soro.2020.0143},
  year = {2021},
  journal = {Soft Robotics}
}

@article{choi_2019_soft,
  author = {Choi, Won Ho and Kim, Sunghwan and Lee, Dongun and Shin, Dongjun},
  month = {07},
  pages = {2539-2546},
  title = {Soft, Multi-DoF, Variable Stiffness Mechanism Using Layer Jamming for Wearable Robots},
  doi = {10.1109/lra.2019.2908493},
  urldate = {2022-09-09},
  volume = {4},
  year = {2019},
  journal = {IEEE Robotics and Automation Letters}
}

@article{hauser_2017_jammjoint,
  author = {Hauser, Simon and Robertson, Matthew J and Auke Jan Ijspeert and Paik, Jamie},
  month = {04},
  pages = {849-855},
  title = {JammJoint: A Variable Stiffness Device Based on Granular Jamming for Wearable Joint Support},
  doi = {10.1109/lra.2017.2655109},
  urldate = {2023-06-30},
  volume = {2},
  year = {2017},
  journal = {IEEE Robotics and Automation Letters}
}

@article{mccabe_2024_combining,
  author = {McCabe, Emily M. and Esser, Daniel S. and Ertop, Tayfun Efe and Kuntz, Alan and Webster III, Robert J.},
  month = {04},
  pages = {711-715},
  publisher = {IEEE},
  title = {Combining Thermoelectrics and Low Melting Point Alloys to Create Reconfigurable Stiff-Compliant Manipulators},
  doi = {10.1109/robosoft60065.2024.10522010},
  url = {https://ieeexplore.ieee.org/document/10522010},
  urldate = {2025-08-06},
  year = {2024},
  journal = {2024 IEEE 7th International Conference on Soft Robotics (RoboSoft)}
}

@article{hwang_2022_shape,
  author = {Hwang, Dohgyu and Barron, Edward J. and Haque, A. B. M. Tahidul and Bartlett, Michael D.},
  month = {02},
  title = {Shape morphing mechanical metamaterials through reversible plasticity},
  doi = {10.1126/scirobotics.abg2171},
  volume = {7},
  year = {2022},
  journal = {Science Robotics}
}

@article{shah_2023_robotic,
  author = {Shah, Dylan S and Woodman, Stephanie J and Buckner, Trevor L and Yang, Ellen J and Kramer-Bottiglio, Rebecca K},
  month = {11},
  pages = {1147-1154},
  publisher = {Institute of Electrical and Electronics Engineers},
  title = {Robotic Skins With Integrated Actuation, Sensing, and Variable Stiffness},
  doi = {10.1109/lra.2023.3337702},
  url = {https://ieeexplore.ieee.org/document/10333261},
  urldate = {2025-08-06},
  volume = {9},
  year = {2023},
  journal = {IEEE Robotics and Automation Letters}
}

@article{pardomuan_2024_vabricbeads,
  author = {Pardomuan, Jefferson and Miyafuji, Shio and Takahashi, Nobuhiro and Koike, Hideki},
  month = {05},
  pages = {1-17},
  publisher = {ACM},
  title = {VabricBeads : Variable Stiffness Structured Fabric using Artificial Muscle in Woven Beads},
  doi = {10.1145/3613904.3642401},
  urldate = {2025-08-06},
  year = {2024},
  journal = {Proceedings of the CHI Conference on Human Factors in Computing Systems}
}

@article{chenal_2014_variable,
  author = {Chenal, Thomas P. and Case, Jennifer C. and Paik, Jamie and Kramer, Rebecca K.},
  month = {09},
  title = {Variable stiffness fabrics with embedded shape memory materials for wearable applications},
  doi = {10.1109/iros.2014.6942950},
  url = {https://ieeexplore.ieee.org/abstract/document/6942950/},
  urldate = {2020-02-03},
  year = {2014},
  journal = {2014 IEEE/RSJ International Conference on Intelligent Robots and Systems}
}

@article{gao_2023_programmable,
  author = {Gao, Weinan and Kang, Jingtian and Wang, Guohui and Ma, Haoxiang and Chen, Xueyan and Kadic, Muamer and Laude, Vincent and Tan, Huifeng and Wang, Yifan},
  month = {10},
  publisher = {Wiley},
  title = {Programmable and Variable‐Stiffness Robotic Skins for Pneumatic Actuation},
  doi = {10.1002/aisy.202300285},
  volume = {5},
  year = {2023},
  journal = {Advanced intelligent systems}
}

@article{song_2025_a,
  author = {Song, Dezhi and Luo, Xiangyu and Yu, Xiangyang and Zhang, Bo and Yang, Zhengbao and Hu, Chengzhi and Shi, Chaoyang},
  month = {01},
  pages = {1-8},
  publisher = {Institute of Electrical and Electronics Engineers},
  title = {A Phase-change-material-based Variable Stiffness Sheath Inspired by a Multi-layer Wave Spring Structure for Flexible Upper Gastrointestinal Endoscopic Robots},
  doi = {10.1109/lra.2025.3568564},
  url = {https://ieeexplore.ieee.org/document/10994675},
  urldate = {2025-08-06},
  year = {2025},
  journal = {IEEE Robotics and Automation Letters}
}

@article{kim_2012_design,
  author = {Kim, Yong-Jae and Cheng, Shanbao and Kim, Sangbae and Iagnemma, Karl},
  month = {10},
  title = {Design of a tubular snake-like manipulator with stiffening capability by layer jamming},
  doi = {10.1109/iros.2012.6385574},
  year = {2012},
  journal = {International Conference on Intelligent Robots and Systems}
}

@article{shamsaalharthy_2024_variable,
  author = {Shamsa Al Harthy and Hadi, M and Wu, Zicong and Seneci, Carlo A and Christos Bergeles},
  month = {06},
  title = {Variable Stiffness Soft Eversion Growing Robot via Temperature Control of Low-Melting Point Alloy Pressurised Medium},
  doi = {10.1109/ISMR63436.2024.10585688},
  url = {https://www.researchgate.net/publication/381225768_Variable_Stiffness_Soft_Eversion_Growing_Robot_via_Temperature_Control_of_Low-Melting_Point_Alloy_Pressurised_Medium},
  year = {2024},
  journal = {IEEE International Symposium on Medical Robotics (ISMR) 2024}
}

@article{narang_2018_mechanically,
  author = {Narang, Yashraj S and Vlassak, Joost J and Howe, Robert D},
  month = {04},
  pages = {1707136-1707136},
  title = {Mechanically Versatile Soft Machines through Laminar Jamming},
  doi = {10.1002/adfm.201707136},
  volume = {28},
  year = {2018},
  journal = {Advanced Functional Materials}
}

@article{brown_2010_universal,
  author = {Brown, E. and Rodenberg, N. and Amend, J. and Mozeika, A. and Steltz, E. and Zakin, M. R. and Lipson, H. and Jaeger, H. M.},
  month = {10},
  pages = {18809-18814},
  title = {Universal robotic gripper based on the jamming of granular material},
  doi = {10.1073/pnas.1003250107},
  volume = {107},
  year = {2010},
  journal = {Proceedings of the National Academy of Sciences}
}

@article{ramachandran_2018_allfabric,
  author = {Ramachandran, Vivek and Shintake, Jun and Floreano, Dario},
  month = {10},
  pages = {1800313},
  title = {All-Fabric Wearable Electroadhesive Clutch},
  doi = {10.1002/admt.201800313},
  urldate = {2020-11-26},
  volume = {4},
  year = {2018},
  journal = {Advanced Materials Technologies}
}

@article{shiva_2016_tendonbased,
  author = {Shiva, Ali and Stilli, Agostino and Noh, Yohan and Faragasso, Angela and Falco, Iris De and Gerboni, Giada and Cianchetti, Matteo and Menciassi, Arianna and Althoefer, Kaspar and Wurdemann, Helge A.},
  month = {07},
  pages = {632-637},
  title = {Tendon-Based Stiffening for a Pneumatically Actuated Soft Manipulator},
  doi = {10.1109/lra.2016.2523120},
  volume = {1},
  year = {2016},
  journal = {IEEE Robotics and Automation Letters}
}

@article{shah_2020_jamming,
  author = {Shah, Dylan S. and Yang, Ellen J. and Yuen, Michelle C. and Huang, Evelyn C. and Kramer‐Bottiglio, Rebecca},
  month = {09},
  pages = {2006915},
  title = {Jamming Skins that Control System Rigidity from the Surface},
  doi = {10.1002/adfm.202006915},
  urldate = {2023-05-09},
  volume = {31},
  year = {2020},
  journal = {Advanced Functional Materials}
}

@article{kier_1985_tongues,
  author = {KIER, WILLIAM M. and SMITH, KATHLEEN K.},
  month = {04},
  pages = {307-324},
  title = {Tongues, tentacles and trunks: the biomechanics of movement in muscular-hydrostats},
  doi = {10.1111/j.1096-3642.1985.tb01178.x},
  url = {https://academic.oup.com/zoolinnean/article-abstract/83/4/307/2648636},
  volume = {83},
  year = {1985},
  journal = {Zoological Journal of the Linnean Society}
}

@article{an_2025_bioinspired,
  author = {An, Jiajun and Zhang, Huayu and Wang, Shengzhi and Li, Zelin and Lin, Han and Zeng, Zihan Oliver and Wen, Qing and Gan, Xingming and Gan, Dongming and Kaur, Upinder and Ma, Xin},
  month = {01},
  pages = {1-8},
  publisher = {Institute of Electrical and Electronics Engineers},
  title = {Bio-inspired Soft Variable-Stiffness Prehensile Tail Enabling Versatile Grasping and Enhancing Dynamic Mobility},
  doi = {10.1109/lra.2025.3559842},
  url = {https://ieeexplore.ieee.org/abstract/document/10960617},
  urldate = {2025-08-06},
  year = {2025},
  journal = {IEEE Robotics and Automation Letters}
}

@article{rus_2015_design,
  author = {Rus, Daniela and Tolley, Michael T.},
  month = {05},
  pages = {467-475},
  title = {Design, fabrication and control of soft robots},
  doi = {10.1038/nature14543},
  url = {https://www.nature.com/articles/nature14543},
  volume = {521},
  year = {2015},
  journal = {Nature}
}

@article{oncayyasa_2023_an,
  author = {Oncay Yasa and Yasunori Toshimitsu and Michelis, Mike Y and Jones, Lewis and Filippi, Miriam and Buchner, Thomas and Katzschmann, Robert K},
  month = {05},
  pages = {1-29},
  publisher = {Annual Reviews},
  title = {An Overview of Soft Robotics},
  doi = {10.1146/annurev-control-062322-100607},
  volume = {6},
  year = {2023},
  journal = {Annual review of control, robotics, and autonomous systems}
}

@article{Shah2021JammingSkins,
  author  = {Shah, Dylan S. and Yang, Ellen J. and Yuen, Michelle C. and Huang, Evelyn C. and Kramer-Bottiglio, Rebecca},
  title   = {Jamming Skins that Control System Rigidity from the Surface},
  journal = {Advanced Functional Materials},
  year    = {2021},
  volume  = {31},
  number  = {1},
  pages   = {2006915},
  doi     = {10.1002/adfm.202006915},
  publisher = {Wiley-VCH}
}

@article{Yang2021TensileJamming,
  author  = {Yang, Bilige and Baines, Robert L. and Shah, Dylan and Patiballa, Sreekalyan and Thomas, Eugene and Venkadesan, Madhusudhan and Kramer-Bottiglio, Rebecca},
  title   = {Reprogrammable soft actuation and shape-shifting via tensile jamming},
  journal = {Science Advances},
  year    = {2021},
  volume  = {7},
  number  = {40},
  pages   = {eabh2073},
  doi     = {10.1126/sciadv.abh2073},
  publisher = {American Association for the Advancement of Science}
}

@article{AlHarthy2025PhaseChangeAlloys,
  author  = {Al Harthy, Shamsa and Sadati, S. M. Hadi and Stockdale, Shannon C. and Neji, Radhouene and Bergeles, Christos},
  title   = {Phase-Change Alloys Enable Localized Reversible Stiffening and Actuation in Steerable Eversion Tip-Growing Robots},
  journal = {Advanced Intelligent Systems},
  year    = {2025},
  pages   = {e202500756},
  doi     = {10.1002/aisy.202500756},
  note    = {Early View},
  publisher = {Wiley-VCH}
}

@article{Hauser2017JammJoint,
  author    = {Hauser, Simon and Robertson, Matthew A. and Ijspeert, Auke Jan and Paik, Jamie},
  title     = {JammJoint: A Variable Stiffness Device Based on Granular Jamming for Wearable Joint Support},
  journal   = {IEEE Robotics and Automation Letters},
  year      = {2017},
  volume    = {2},
  number    = {2},
  pages     = {849--855},
  doi       = {10.1109/LRA.2017.2655109}
}
\end{document}